%% file: main.tex
\documentclass{article}

\usepackage{soul_style}
\usepackage[export]{adjustbox}
\usepackage[utf8]{inputenc} %
\usepackage[T1]{fontenc}    %
\usepackage{newunicodechar}
\usepackage{hyperref}       %
\usepackage{xcolor}
\usepackage[normalem]{ulem} %
\hypersetup{
    colorlinks=true,      %
    linkcolor=blue,      %
    urlcolor=blue,       %
    citecolor=blue,      %
    linkbordercolor=blue, %
    urlbordercolor=blue,
    citebordercolor=blue,
    pdfborderstyle={/S/U/W 1}, %
}
\usepackage[utf8]{inputenc}
\usepackage{booktabs} 
\usepackage{tabularx} 
\usepackage{pifont}   
\usepackage{xcolor}   

\newcommand{\xmark}{\ding{55}} 
\usepackage{float}
\usepackage{url}            %
\usepackage{booktabs}       %
\usepackage{amsfonts}       %
\usepackage{nicefrac}       %
\usepackage{microtype}      %
\usepackage{lipsum}		%
\usepackage{graphicx}
\usepackage{natbib}
\usepackage{doi}
\usepackage{amsmath}
\usepackage{amssymb} %
\usepackage{xspace}
\usepackage{enumitem}
\usepackage{multirow}
\usepackage{subcaption} 
\usepackage{makecell}
\usepackage{hyperref, cleveref}
\usepackage{pifont}
\usepackage[inkscapelatex=false]{svg}
\usepackage{caption}
\usepackage{wrapfig}
\usepackage{algorithm}
\usepackage{algorithmic}
\usepackage{threeparttable}
\usepackage{overpic}

\setlist[itemize]{leftmargin=*}
\setlist[enumerate]{leftmargin=*}
\setlist[description]{leftmargin=*}

\newcommand{\soulxflashhead}{SoulX-FlashHead\xspace}

\definecolor{midnightgreen}{rgb}{0.0, 0.29, 0.33}

\title{\textnormal{\soulxflashhead: Oracle-guided Generation of Infinite Real-time Streaming Talking Heads}}
\author{
\normalfont Tan Yu\thanks{Equal contribution. \{qiaoqian,yutan\}@soulapp.cn, Core Contributors: Tan Yu, Qian Qiao, Le Shen, Ke Zhou}, Qian Qiao\textsuperscript{*,}$^\dag$, Le Shen\textsuperscript{*}, \normalfont Ke Zhou, Jincheng Hu, Dian Sheng, \\Bo Hu, Haoming Qin,
Jun Gao, Changhai Zhou,
 Shunshun Yin, Siyuan Liu\thanks{Corresponding authors. Project Leader: liusiyuan@soulapp.cn}   \\
    AIGC Team, Soul AI Lab, China \\
      \textbf{Project Page:} \href{https://soul-ailab.github.io/soulx-flashhead/}{https://soul-ailab.github.io/soulx-flashhead/}\\
}
\renewcommand{\headeright}{\raisebox{-0.2\height}{\includegraphics[height=1.8em]{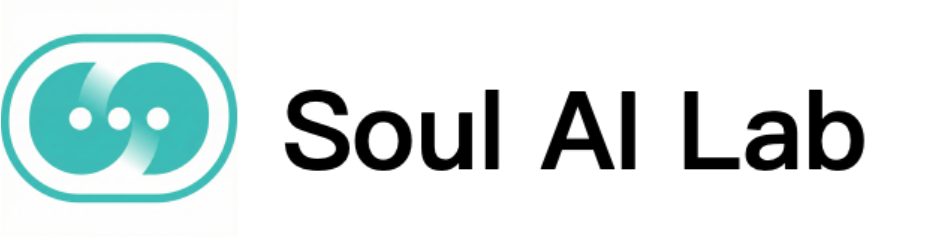}}}
\renewcommand{\shorttitle}{SoulX-FlashHead Technical Report}

\usepackage{amsmath}
\usepackage{CJKutf8}
\begin{document}
\maketitle

\input{content_arxiv/0.abstract}
\input{content_arxiv/1.Introduction}
\input{content_arxiv/2.Data}

\input{content_arxiv/3.Method}
\input{content_arxiv/4.Experiments}
\input{content_arxiv/5.Conclusion}
\input{content_arxiv/6.Statement}
\bibliographystyle{unsrtnat}
\bibliography{references}  

\clearpage
\appendix

\end{document}

%% file: content_arxiv/0.abstract.tex
\begin{abstract}
Achieving a balance between high-fidelity visual quality and low-latency streaming remains a formidable challenge in audio-driven portrait generation. Existing large-scale models often suffer from prohibitive computational costs, while lightweight alternatives typically compromise on holistic facial representations and temporal stability. In this paper, we propose SoulX-FlashHead, a unified 1.3B-parameter framework designed for real-time, infinite-length, and high-fidelity streaming video generation. To address the instability of audio features in streaming scenarios, we introduce Streaming-Aware Spatiotemporal Pre-training equipped with a Temporal Audio Context Cache mechanism, which ensures robust feature extraction from short audio fragments. Furthermore, to mitigate the error accumulation and identity drift inherent in long-sequence autoregressive generation, we propose Oracle-Guided Bidirectional Distillation, leveraging ground-truth motion priors to provide precise physical guidance. We also present VividHead, a large-scale, high-quality dataset containing 782 hours of strictly aligned footage to support robust training. Extensive experiments demonstrate that SoulX-FlashHead achieves state-of-the-art performance on HDTF and VFHQ benchmarks. Notably, our Lite variant achieves an inference speed of 96 FPS on a single NVIDIA RTX 4090, facilitating ultra-fast interaction without sacrificing visual coherence.
\end{abstract}

%% file: content_arxiv/1.Introduction.tex
\section{Introduction}
In the realm of real-time audio-driven portrait generation, large-scale Diffusion Transformer models~\citep{gao2025wan,zhong2025anytalker,guo2024liveportrait,yang2025infinitetalk} have demonstrated exceptional performance in high-fidelity streaming generation. However, their deployment poses significant challenges. Approaches such as LiveAvatar~\citep{huang2025live} and SoulX-FlashTalk~\citep{shen2025soulx} are hindered by heavy computational overhead and complex pipeline parallelism. This renders low-latency streaming interaction on consumer-grade GPUs largely infeasible. Conversely, traditional quantization and pruning techniques often degrade video fidelity and struggle to capture complex facial micro-expressions or precise lip synchronization.

As shown in Tab.~\ref{table_diff_model_comparison}, we compare existing methodologies along four dimensions including streaming capability, real-time inference, infinite-length generation, and holistic representation. We define Holistic Representation as the modeling of pixel-level latent spaces utilizing image or video VAEs. This approach ensures the preservation of complete facial textures and visual coherence. In contrast, approaches like Ditto~\citep{li2025ditto} and SadTalker~\citep{zhang2023sadtalker} rely on motion VAE–based representations. These are inherently more abstract and lack a unified description of the facial structure. In contrast, approaches like Ditto~\citep{li2025ditto} and SadTalker~\citep{zhang2023sadtalker} rely on motion VAE–based representations which are inherently abstract and lack a unified facial representation. A clear trade-off exists between efficiency and representational capacity. Lightweight models enable real-time streaming but lack holistic details while high-fidelity models like Hallo3~\citep{cui2025hallo3} and AniPortrait~\citep{wei2024aniportrait} typically fail to support real-time or streaming inference. A unified framework satisfying all requirements within a moderate parameter scale remains elusive.

To bridge this gap, we introduce SoulX-FlashHead. This is a 1.3B-parameter model designed for real-time streaming video generation. Unlike previous works that compromise between speed and quality, SoulX-FlashHead achieves a balance by employing a two-stage training scheme comprising Streaming-Aware Spatiotemporal Pre-training and Oracle-Guided Bidirectional Distillation. This design specifically addresses two fundamental challenges in the field.

Data Noise and Audio Feature Instability. Lightweight models often struggle to learn precise conditional mappings from noisy datasets. Streaming interaction necessitates processing extremely short audio fragments which are typically around 1.32 seconds. Traditional self-supervised audio models like Wav2Vec~\citep{baevski2020wav2vec} are prone to feature space distribution shifts or collapse when handling such short sequences. This leads to signal distortion. We address this via Streaming-Aware Spatiotemporal Pre-training. At the data level, we constructed a rigorous cleaning pipeline to refine VividHead which is a high-quality dataset containing over 10,000 hours of highly aligned data. Furthermore, we introduced a Temporal Audio Context Cache mechanism during pre-training. This explicitly caches historical audio to compensate for context deficiency in short-slice inputs and ensures robust audio representations and precise audio-visual alignment.

Error Accumulation in Long-Sequence Generation. Real-time streaming requires autoregressive prediction where minor deviations in 1.3B models amplify rapidly and cause facial distortion or identity drift. While distribution distillation (DMD)~\citep{yin2024one} is commonly used to reduce error accumulation, existing methods ignore the misalignment between pre-training and distillation phases. Specifically, the teacher often uses ground truth motion frames while the student relies on predictions. This results in inaccurate guidance. We propose Oracle-Guided Bidirectional Distillation to overcome this. By utilizing Ground Truth motion frames as "Oracle" conditional anchors, we provide clear physical priors. This strong constraint mechanism suppresses error diffusion in long sequences and fully exploits the teacher model to improve fidelity at low inference steps.

SoulX-FlashHead is provided in two flexible deployment versions. SoulX-FlashHead-Lite targets ultra-fast interaction scenarios and achieves real-time inference on a single RTX 4090 GPU. SoulX-FlashHead-Pro pursues superior detail and enables real-time generation on dual RTX 5090 GPUs.

\begin{table}[htbp]
\centering
\caption{Comparison of different audio-driven portrait generation methods. Our model uniquely achieves a balance of streaming capability, real-time performance, infinite generation length, and holistic representation with a lightweight 1.3B parameter size.}
\label{table_diff_model_comparison}
\begin{tabular}{l|cccc|l}
\toprule
\textbf{Method} & Stream & Real-Time & Inf-len & Holistic Representation & \textbf{Size} \\
\midrule
SadTalker~\citep{zhang2023sadtalker}     & \checkmark & \checkmark & \checkmark & \xmark & 0.2B \\
Aniportrait~\citep{wei2024aniportrait}   & \xmark & \xmark & \checkmark & \checkmark & 1.7B \\
EchoMimicV3~\citep{chen2025echomimic}   & \checkmark & \xmark & \xmark & \checkmark & 1.3B \\
Ditto~\citep{li2024ditto}         & \checkmark & \checkmark & \checkmark & \xmark & 0.2B \\
Hallo3~\citep{cui2025hallo3}        & \checkmark & \xmark & \xmark & \checkmark & 5B \\
Sonic~\citep{ji2025sonic}         & \xmark & \xmark & \checkmark & \checkmark & 1.5B \\ 
\hline
\textbf{SoulX-FlashHead} & \checkmark & \checkmark & \checkmark & \checkmark & \textbf{1.3B} \\ 
\bottomrule
\end{tabular}
\end{table}

%% file: content_arxiv/2.Data.tex
\section{Data}
To enable high-fidelity and vivid portrait video generation, we constructed VividHead~\footnote{https://huggingface.co/datasets/Soul-AILab/VividHead}, a large-scale and high-quality dataset. Recognizing that data quality dictates the upper bound of downstream model performance, we developed the comprehensive data processing pipeline shown in Fig.~\ref{figure_data_pipe} to transform unstructured raw web videos into clean and semantically rich training data. This process comprises two core phases where Data Preprocessing handles acquisition and standardization while Data Filtering and Annotation performs strict screening and fine-grained labeling.

VividHead consists of 330,000 high-quality short clips ranging from 3 to 60 seconds with a total duration of 782 hours. Each sample features $512\times 512$ resolution image sequences, strictly time-aligned speech audio, and rich metadata encompassing language, ethnicity, and age. We restricted the collection to samples containing a single visible speaker with an active head region.

Tab.~\ref{table_dataset_comparison} compares VividHead with existing mainstream datasets. Distinct from collections limited to laboratory settings such as MEAD or lower resolutions like HDTF, VividHead balances in-the-wild diversity with high visual quality standards across 15 languages and diverse demographics.

\input{new_figs/figure_data_pipe}
\input{new_tables/table_dataset_comparison}

\subsection{Data Processing Pipeline}
\subsubsection{Data Preprocessing stage}

We constructed the initial data pool by aggregating public datasets~\citep{ephrat2018looking,zhang2021flow,cui2025hallo3,chen2025talkvidlargescalediversifieddataset,zhang2025speakervid} and extensive web resources. To address the redundancy inherent in large-scale multi-source collection, we applied a verification mechanism based on source video IDs and MD5 hash values to eliminate duplicate content and guarantee sample uniqueness.

For video segmentation, we employed distinct strategies where public datasets with existing timestamps were cropped precisely while unlabelled web data underwent adaptive scene detection via PySceneDetect~\citep{Castellano_PySceneDetect_2024}. This approach divided long videos into coherent clips ranging from 5 to 50 seconds to preserve semantic context. All clips were subsequently normalized to a unified frame rate of 25 fps using FFMPEG to ensure temporal consistency for downstream modeling.

\subsubsection{Data filter and annotation stage}

In the data filtering phase, we first perform face detection and mask extraction on the initial frame of each clip, and valid clips are finally cropped to a resolution of $512\times 512$ pixels centered on the detected face. Subsequently, optical flow analysis~\citep{dosovitskiy2015flownet} is employed to identify and remove sequences containing abrupt scene transitions. To ensure consistent facial visibility, we inspect raw footage at a sampling rate of one frame per second and exclude videos where the proportion of frames lacking a detectable face exceeds a predefined threshold. We further utilize DWpose~\citep{yang2023effective} to extract body keypoints and strictly filter out clips exhibiting hand-over-face occlusion to prevent generation artifacts. Finally, we employ the SyncNet~\citep{chung2016out} model to calculate LSE-C and LSE-D confidence scores to discard samples with poor audio-visual alignment.

High-quality clips passing these rigorous filters proceed to the automated data annotation and feature extraction pipeline. High-precision detectors extract face masks at the visual level to decouple the foreground from complex backgrounds. For audio, we separate streams and employ pre-trained Wav2Vec models to extract streaming audio embeddings as robust driving features. We also annotate clips with multi-dimensional attributes including gender, age, ethnicity, and language to enhance the capabilities of the model in controllable generation tasks.

%% file: new_figs/figure_data_pipe.tex
\begin{figure*}[h]
    \centering
    \includegraphics[width=1.0\linewidth]{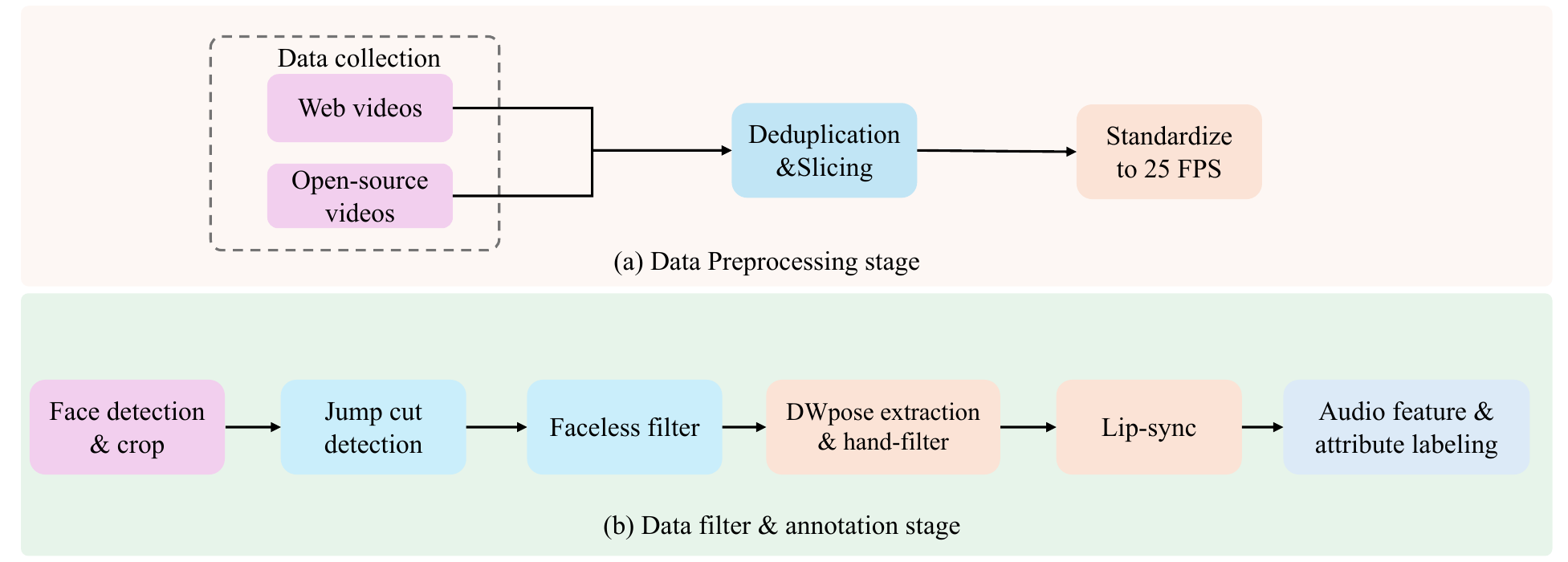}
    \caption{Overview of our comprehensive data filtering pipeline. We obtain 782 hours of high-quality audio–video data from 10k hours
    }
    \label{figure_data_pipe}
\end{figure*}

%% file: new_tables/table_dataset_comparison.tex
\begin{table}[htbp]
  \centering
  \caption{Comparison of TalkVivid with existing talking head datasets. TalkVivid distinguishes itself through a combination of large scale, high resolution, and diverse attribute coverage in wild scenarios.}
  \label{table_dataset_comparison}
  \small 
  \resizebox{\linewidth}{!}{%
  \begin{tabular}{lccccccccc}
    \toprule
    Dataset & Speakers & Face Crop & Clips & Hours & Resolution & Language & Age & Ethnicty & Source \\
    \midrule
    MEAD~\citep{kaisiyuan2020mead} & 60 & \checkmark & 281.4K & 39 & 384p & English & 20-35 & - & Lab \\
    HDTF~\citep{zhang2021flow} & 362 & \checkmark & 10K & 15.8 & 512p & - & - & - & wild \\
    AVspeech~\citep{ephrat2018looking} & 150k & \ding{55} & 2.5M & 4700 & 720p, 1080p & - & - & - & wild \\
    Hallo3~\citep{cui2025hallo3} & - & \checkmark & 101.5K & 70 & 720p & - & - & - & wild \\
    OpenHumanVid~\citep{zheng2024open} & - & \ding{55} & 13.4M & 16.7k & 720p & - & - & - & wild \\
    TalkVid~\citep{chen2025talkvidlargescalediversifieddataset} & 7729 & \ding{55} & 281.4K & 1244 & 1080p, 2160p & 15 langs & 0-60+ & 3 & wild \\
    SpeakerVid~\citep{zhang2025speakervid} & 83k & \ding{55} & 5.2M & 8.7K & 1080p & - & - & - & wild \\
    \midrule
    \textbf{VividHead} & \textbf{60k} & \checkmark & \textbf{330K} & \textbf{782} & \textbf{512p} & \textbf{15 langs} & \textbf{0-60+} & \textbf{3} & \textbf{wild} \\
    \bottomrule
  \end{tabular}}
\end{table}

%% file: content_arxiv/3.Method.tex
\section{SoulX-FlashHead}
\input{new_figs/figure_pipe}
We present SoulX-FlashHead, a framework designed to generate audio-synchronized videos that faithfully preserve reference image content given audio and image inputs. As illustrated in Figure~\ref{figure_pipe}, the architecture builds upon a 1.3B-parameter DiT backbone and employs a two-stage training strategy comprising Streaming-Aware Spatiotemporal Pre-training and Oracle-Guided Bidirectional Distillation. This training paradigm is supported by a comprehensive real-time inference acceleration pipeline.

\subsection {Model Architecture}
\noindent \textbf{3D-VAE.}
To strictly address the trade-off between visual fidelity and inference latency, we employ distinct VAE architectures for our Pro and Lite variants. The Pro model utilizes the WAN 2.1 VAE~\citep{wan2025wan} which encodes video frames into compact latent representations with a spatio-temporal downsampling factor of $4 \times 8 \times 8$. This configuration prioritizes high-fidelity reconstruction and fine-grained detail preservation suitable for quality-critical scenarios. Conversely, the Lite model incorporates the LTX-VAE~\citep{hacohen2024ltx} to optimize for real-time performance. LTX-VAE achieves aggressive compression by downsampling inputs by a factor of $(32, 32, 8)$, yielding a pixel-to-token ratio of $8192:1$—approximately $32\times$ higher than the WAN scheme. While LTX-VAE integrates a diffusion decoding step to mitigate detail loss inherent to such high compression, it fundamentally serves as the engine for speed-balanced inference in latency-sensitive applications.

\noindent \textbf{Diffusion Transformer (DiT).} DiT~\citep{peebles2023scalable} replaces the U-Net~\citep{ronneberger2015u} in latent diffusion models (LDMs)~\citep{rombach2022high} with a Transformer, enhancing capacity and scalability across space and time. In video generation, it is often combined with a causal 3D VAE~\citep{kingma2013auto,zheng2024open} for spatio-temporal compression, while conditional inputs are incorporated via adaptive normalization or cross-attention. The objective is to learn a vector field by minimizing the mean squared error (MSE) between predicted and ground-truth velocities:
\begin{equation}
\mathcal{L}_{\text{FM}}(\theta) = \mathbb{E}_{t,\, p_{t}(\mathbf{x})} \| \mathbf{v}_t - \mathbf{u}_t \|^2,
\end{equation}
where $\mathbf{x}_t = t \cdot \mathbf{x}_1 + (1 - t) \cdot \mathbf{x}_0$. We use Wan2.1 T2V~\citep{wan2025wan} (1.3B) as our baseline, balancing performance and efficiency.

\noindent \textbf{Audio Condition injection.}
To facilitate Streaming-Aware Spatiotemporal Pre-training and address streaming input instability, we designed a Temporal Audio Context Cache mechanism using a fixed-length window. We define the audio context window length as $T_{max} = 8\text{s}$. For any given raw audio stream $\mathcal{A}_{raw}$, we construct a standardized input queue $\mathcal{Q}_{audio}$ as follows:
\begin{equation}
\label{eq_2}
\mathcal{Q}_{audio} = \begin{cases}
\text{Concat}(\mathbf{0}_{\text{pad}}, \mathcal{A}_{raw}), & \text{if } |\mathcal{A}_{raw}| < T_{max} \\
\mathcal{A}_{raw}[-T_{max}:], & \text{if } |\mathcal{A}_{raw}| \ge T_{max}
\end{cases}
\end{equation}
Here, $\mathbf{0}_{\text{pad}}$ denotes silence audio padding, and $\mathcal{A}_{raw}[-T_{max}:]$ represents the trailing $8\text{s}$ segment of the audio stream. We then encode $\mathcal{Q}_{audio}$ using Wav2Vec. To integrate multi-level semantic cues, we aggregate multi-layer Wav2Vec features, yielding the full audio feature sequence $E_{audio} \in \mathbb{R}^{S \times Layers\times D}$, where $S$ corresponds to the feature frame count for the $8\text{s}$ duration. During training and inference, we ensure precise synchronization with the current video frame by extracting the last $N$ time-aligned audio feature frames from $E_{audio}$ as the driving condition:
\begin{equation}
    Z_{cond} = E_{audio}[-N:]
\end{equation}
Finally, $Z_{cond}$ is injected into selected DiT blocks via pixel-wise cross-attention to drive facial motion generation in an end-to-end manner.

\noindent \textbf{Image Conditioning and Anti-Drift.} Image conditioning is critical for maintaining visual consistency in long-video generation. To reinforce identity preservation, we forgo complex attention injection mechanisms in favor of channel-wise concatenation of the reference image latent with the input noise. This direct injection strategy provides a robust spatial structural prior, ensuring that high-frequency details from the reference image maintain precise pixel-level alignment throughout generation. Consequently, this approach effectively mitigates identity drift in long sequences.

\noindent \textbf{Context History and First-Frame Adaptation.} To ensure smooth transitions between video chunks, we utilize motion frames as context. However, streaming inference presents a unique "cold start" challenge where the first chunk lacks historical video context. We address this via a Dynamic Motion Frame Sampling strategy during pre-training. Specifically, we sample motion frames from the beginning of the Ground Truth video: with a probability of 0.9, we use the first $n$ frames, and with a probability of 0.1, we use only the single first frame. This distribution effectively simulates the initial inference phase where the motion context consists solely of the reference image, enabling the model to adapt seamlessly to the first chunk's generation.

\subsection{Model Training}
To achieve high-fidelity real-time streaming generation with a 1.3B-parameter architecture, we employ a two-stage training strategy comprising Streaming-Aware Spatiotemporal Pre-training and Oracle-Guided Bidirectional Distillation. This approach addresses the instability of audio features in streaming environments and the accumulation of errors during long-sequence autoregressive generation.

\textbf{Stage 1: Streaming-Aware Spatiotemporal Pre-training}

Streaming talking head generation faces an inherent conflict between the need for high-fidelity lip synchronization and the instability of real-time audio feature extraction. In live inference scenarios, the model must process extremely short audio fragments of approximately 1.32 seconds. This fragmentation often induces distribution shifts or feature collapse within traditional self-supervised frameworks and results in signal distortion. To mitigate these issues at the data level, we established a rigorous processing pipeline that refines over 10,000 hours of raw footage into TalkVivid which is a high-quality dataset containing 1,000 hours of strictly aligned data. This foundation ensures the model learns from a robust and clean data distribution.

To guarantee stable feature extraction under streaming constraints, we introduce the Temporal Audio Context Cache mechanism as formulated in Equation 2. We maintain a fixed-length audio window of 8 seconds where raw audio inputs are padded with silence or truncated to ensure consistent feature extraction and context integrity. Furthermore, we adopt a probabilistic motion conditioning strategy to adapt the model to both continuous streaming and initial generation phases. During training, we utilize $n$ Ground Truth frames as motion context with a probability of 0.9 to capture temporal dependencies. Conversely, we use a single frame with a probability of 0.1 to simulate the cold start scenario where only the reference image is available.

\textbf{Stage 2: Oracle-Guided Bidirectional Distillation}

Real-time streaming inference necessitates autoregressive generation where the model predicts future frames based on its own history. This process often suffers from error accumulation that leads to severe identity drift. To enable real-time performance, we initially adopt the Distribution Matching Distillation (DMD) framework to compress sampling steps and eliminate the need for classifier-free guidance. DMD optimizes the Kullback–Leibler divergence to minimize the distributional discrepancy between the original teacher and the distilled student at each timestep $t$. The gradient update is formulated as:
\begin{equation}
\label{eq:dmd}
    \nabla_\theta \mathcal{L}_{\text{DMD}} = -\mathbb{E}_{t, \mathbf{z}} \left[ \left( s_{\text{real}}(\psi(G_\theta(z),t), t) - s_{\text{fake}}(\psi(G_\theta(z),t), t) \right) \frac{\partial G_\theta(\mathbf{z})}{\partial \theta} \right],
\end{equation}
Here, $s_{\text{real}}$ represents the frozen teacher score network modeling the true distribution, while $s_{\text{fake}}$ tracks the evolving student distribution generated by $G_\theta$. All components are initialized from the Stage-1 Pretrained model.

Standard DMD overlooks the temporal dependency in streaming where the student relies on imperfect historical context. To address this, we introduce Oracle-Guided Bidirectional Distillation. Inspired by the error correction mechanism in Self-Forcing++, we explicitly simulate the autoregressive inference process during training. The student generator synthesizes $K$ consecutive video chunks where each chunk is conditioned on its previously generated motion frames $\mathbf{m}_{pred}$ rather than ground truth.

We implement a Stochastic Truncation Strategy to balance computational efficiency with training stability. Instead of unrolling the full computation graph for $K$ chunks, we randomly sample a truncation length $k$ from a uniform distribution and generate only the first $k$ chunks. During backpropagation, gradients are retained only for a randomly sampled denoising step $t'$ of the $k$-th chunk, while strictly detaching all preceding steps from the computational graph.

Crucially, we leverage an Oracle supervision signal where the teacher model remains conditioned on the ground truth motion history $\mathbf{m}_{gt}$. This contrasts with the student which is conditioned on its accumulated history $\mathbf{m}_{pred}$. We further enforce trajectory alignment by imposing a latent space regression loss. The final objective aggregates the distribution matching loss and the regression penalty:
\begin{equation}
    \label{eq:stochastic_truncation}
    \mathcal{L}_\text{total} = \mathbb{E}_{k, t'} \left[ 
    \underbrace{D_\text{KL} \left( \underbrace{p_{\phi}(\mathbf{x}_0 | \mathbf{m}_\text{gt})}_{s_{\text{real}}} \, \Big\| \, \underbrace{p_{\theta}(\mathbf{x}_0 | \mathbf{m}_\text{pred})}_{s_{\text{fake}}} \right)}_{\mathcal{L}_{\text{DMD}}} 
    + \lambda 
    \underbrace{\| \mathbf{z}_\text{student}^{(k)} - \mathbf{z}_\text{gt}^{(k)} \|_2^2}_{\mathcal{L}_{\text{reg}}} 
    \right]
\end{equation}
The $\mathcal{L}{\text{DMD}}$ term utilizes the Oracle distribution $p_{\phi}$ to guide the drifting student distribution $p_{\theta}$ back to the optimal manifold. Simultaneously, $\mathcal{L}_{\text{reg}}$ minimizes the Euclidean distance between the student's latent output and the ground truth to ensure physical trajectory alignment.

\subsection{Real-time Inference Acceleration}

To enable low-latency inference on consumer-grade hardware (e.g., NVIDIA RTX 4090 and RTX 5090), we implement a full-stack acceleration pipeline tailored for the 1.3B-parameter model.

\noindent \textbf{Hybrid Sequence Parallelism.} The primary computational cost lies in the DiT attention layers. We employ Hybrid Sequence Parallelism via xDiT to distribute the attention workload. By combining Ulysses and Ring Attention mechanisms, we achieve significant speedups in single-step inference for multi-GPU setups compared to standard implementations.

\noindent \textbf{Kernel Optimization.} We adopt FlashAttention-2 to optimize attention operations at the kernel level. This implementation maximizes memory bandwidth utilization on NVIDIA Ada Lovelace and Blackwell architectures. By minimizing memory access overhead and optimizing IO complexity, FlashAttention-2~\citep{dao2023flashattention2} further reduces attention latency.

\noindent \textbf{Runtime Optimization.} Finally, we utilize \texttt{torch.compile} to unify the inference pipeline. This eliminates Python runtime overhead and enables graph-level fusion, ensuring optimal memory usage and execution efficiency on the target hardware.

%% file: new_figs/figure_pipe.tex
\begin{figure*}[h]
    \centering
    \includegraphics[width=1.0\linewidth]{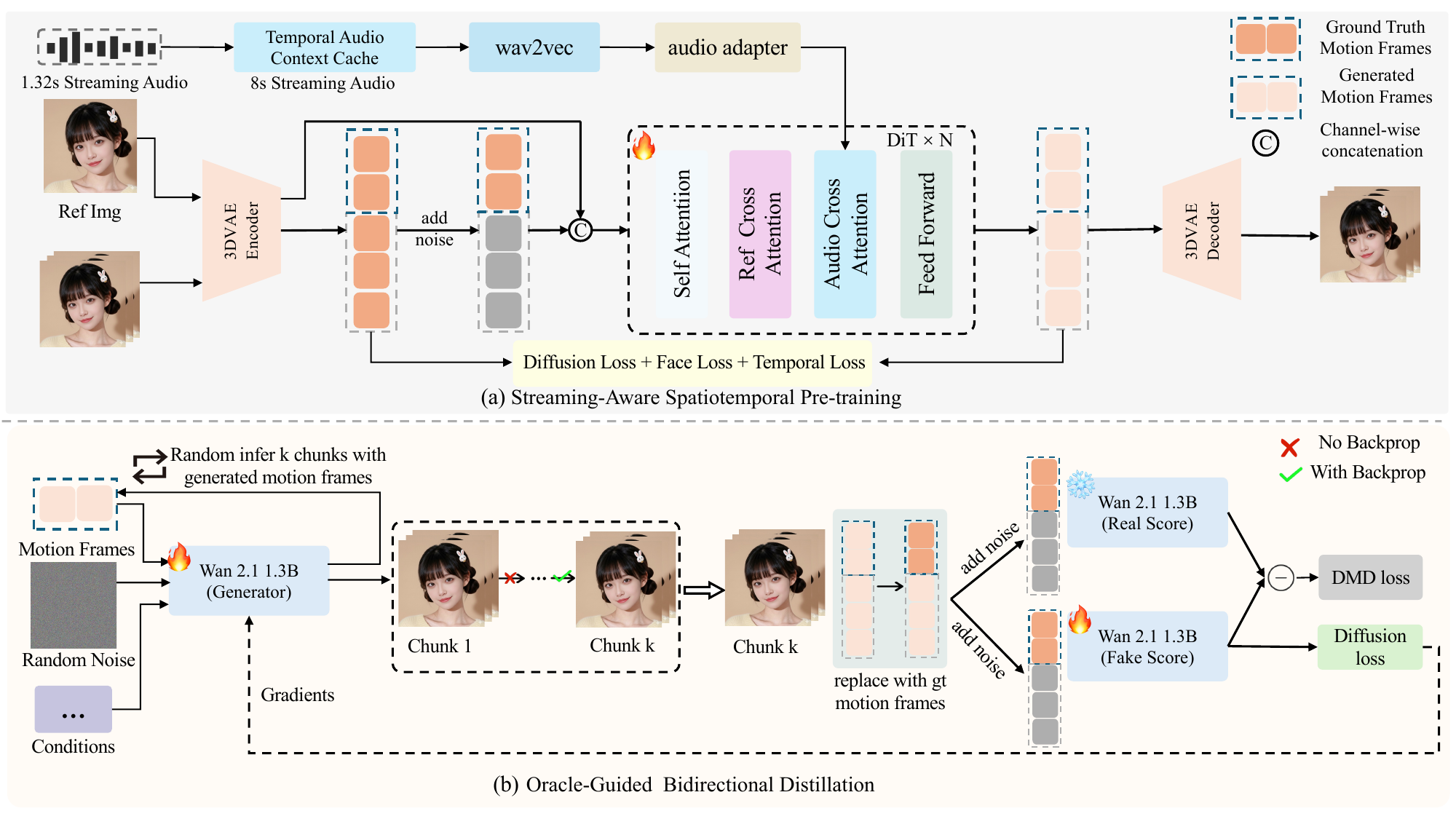}
    \caption{Framework Overview of SoulX-FlashHead. (a) Stage 1: Streaming-Aware Spatiotemporal Pre-training. We employ a Temporal Audio Context Cache to stabilize feature extraction from short streaming audio and utilize channel-wise concatenation for robust reference image injection. (b) Stage 2: Oracle-Guided Bidirectional Distillation. To mitigate error accumulation, the Student generates autoregressively conditioned on its own historical predictions, while the Teacher utilizes Ground Truth motion frames as an "Oracle" guide. The model is optimized via a Stochastic Truncation Strategy using DMD and latent regression losses.
    }
    \label{figure_pipe}
\end{figure*}

%% file: content_arxiv/4.Experiments.tex
\section{Experiments}
\noindent \textbf{Implementation Details.} We build our model upon the Wan2.1 T2V (1.3B) architecture, optimizing it to satisfy real-time constraints. Stage 1 pretrains at $2 \times 10^{-4}$ higher learning rate, incorporating a warm-up strategy and optimized using the AdamW optimizer. The model is trained using 32 NVIDIA H20 GPUs. In the first stage, the global batch size is set to 256 and the model is trained for $100,000$ steps. For the subsequent distillation stage, We adhere to the Self-Forcing training paradigm, setting learning rates to $2 \times 10^{-6}$ for the Generator and $4 \times 10^{-7}$ for the Fake Score Network with a 1:5 update ratio. To simulate error accumulation in long-horizon generation, the Generator synthesizes up to $K=5$ consecutive chunks during distillation. To accommodate variable aspect ratios in real-world data, we employ a bucketing strategy across both SFT and distillation stages. All experiments utilize a cluster of 32 NVIDIA H20 GPUs with a per-GPU batch size of 1.

\noindent \textbf{Evaluation Metrics.} We sampled 75 videos from the HDTF and VFHQ datasets for evaluation and compared our method against state-of-the-art approaches including SadTalker~\citep{zhang2023sadtalker}, Aniportrait~\citep{wei2024aniportrait}, EchoMimic~\citep{chen2025echomimic}, Ditto~\citep{li2024ditto}, and Hallo3~\citep{cui2025hallo3}. Our assessment relies on multiple metrics to provide a comprehensive analysis. We use the Fréchet Inception Distance~\citep{heusel2017gans} to measure distributional discrepancies between generated and real frames and the Fréchet Video Distance~\citep{unterthiner2019fvd} to capture temporal consistency. To assess audio-visual synchronization accuracy and smoothness, we incorporate Sync-C~\citep{chung2016out} for lip motion alignment and Sync-D for the temporal stability of lip dynamics. Finally, we report inference efficiency in frames per second measured on a single NVIDIA H20 GPU.

\subsection{Performance of SoulX-FlashHead}
\input{new_tables/tabble_hdtf_results}
\input{new_tables/table_vfhq_results}
\noindent \textbf{Main Results.}
We performed a comprehensive quantitative comparison with state-of-the-art (SOTA) methods on two benchmark datasets: HDTF~\citep{zhang2021flow} and VFHQ~\citep{xie2022vfhq}. The quantitative results are summarized in Tab.~\ref{tabble_hdtf_results} and Tab.~\ref{table_vfhq_results}. On the HDTF dataset, the non-streaming Pro model achieves an FID of 8.31 and outperforms EchoMimic. In the streaming setting, the Pro model yields an FID of 9.97 which surpasses diffusion baselines including Hallo3 and AniPortrait. Regarding video smoothness measured by FVD, the Pro model scores 103.14 in non-streaming and 111.38 in streaming modes. Both results improve upon Sonic which scores 113.31, indicating that our method generates videos with high temporal coherence. Experiments on the VFHQ dataset demonstrate significant advantages in lip synchronization. The Pro model achieves Sync-C scores of 5.60 for non-streaming and 5.53 for streaming. These scores exceed both SadTalker at 4.49 and Sonic at 4.64. These results validate the effectiveness of the Oracle-Guided Bidirectional Distillation strategy in aligning lip movements with audio signals and maintaining synchronization precision in complex natural scenarios. Regarding inference efficiency and streaming strategy analysis, the Lite model achieves an inference speed of 96 FPS. Compared to the lightweight method Ditto, the Lite model maintains real-time performance while delivering superior visual quality. Against computationally intensive models such as Hallo3 and EchoMimic, our approach offers a substantial speed advantage. A comparison between streaming and non-streaming data reveals minimal performance degradation. For instance, on HDTF, the streaming Lite model retains a Sync-C score of 4.21 and a low FVD. This confirms that the Temporal Audio Context Cache effectively mitigates audio feature collapse during real-time streaming while the motion frame sampling strategy successfully alleviates cold-start issues.

In conclusion, the Pro model achieves state-of-the-art visual and synchronization quality, while the Lite model strikes an optimal balance between high fidelity and real-time responsiveness, outperforming existing methods in comprehensive capability.

\input{new_figs/figure_qua}
\noindent \textbf{Long Video Generation.} We further evaluated the model on the challenging 60-second long video generation task\footnote{The original videos for these comparisons can be found on our project page.} task at 25 fps against state-of-the-art methods. at 25 fps against state-of-the-art methods. Fig.~\ref{figure_qua} demonstrates that our model maintains high-fidelity generation capabilities throughout the entire duration. The green indicators highlight severe error accumulation and artifacts in Hallo3 whereas our method ensures structural stability. In terms of lip synchronization marked by yellow boxes, methods relying on abstract motion representations like Ditto and SadTalker exhibit noticeable desynchronization while our approach maintains precise audio-visual alignment. Furthermore, the red regions reveal that SadTalker fails to preserve holistic consistency where structural connections between the headgear and the subject break due to the absence of unified pixel-level modeling. Our method conversely preserves robust holistic integrity.

%% file: new_tables/tabble_hdtf_results.tex
\begin{table}[t]
\centering
\caption{\textbf{Quantitative comparison on the HDTF dataset.} Models marked with $^*$ support streaming, while those marked with $^\triangle$ do not.  \textbf{For non-streaming methods, FPS is not a meaningful metric.}
}
\label{tabble_hdtf_results}
\small 
\setlength{\tabcolsep}{10pt} 
\begin{tabular}{lccccc}
\toprule
\textbf{Method} & \textbf{FID$\downarrow$} & \textbf{FVD$\downarrow$} & \textbf{Sync-C$\uparrow$} & \textbf{Sync-D$\downarrow$} & \textbf{FPS$\uparrow$} \\
\midrule
SadTalker$^*$~\citep{zhang2023sadtalker}    & 21.58 & 207.67 & 4.60 & 9.21  & 2.17 \\
Aniportrait$^\triangle$~\citep{wei2024aniportrait}  & 19.83 & 242.29 & 1.89 & 11.91 & - \\
EchoMimic$^*$~\citep{chen2025echomimic}    & 9.00  & 155.71 & 3.56 & 10.22 & 0.81 \\
Ditto$^*$~\citep{li2024ditto}        & 12.35 & 199.13 & 3.57 & 10.49  & \underline{45.04} \\
Hallo3$^*$~\citep{cui2025hallo3}       & 15.95 & 160.94 & 3.18 & 10.72 & 0.16 \\
Sonic$^\triangle$~\citep{ji2025sonic}        & 13.53 & 113.31 & 5.17 & 8.69 & - \\
\midrule
SoulX-FlashHead  (Lite)$^*$         & 11.37 & 126.52 & 4.21 & 9.49  & \textbf{96}\\
SoulX-FlashHead  (Pro)$^*$       & \underline{9.97} & \underline{111.38} & \underline{5.73} & \underline{8.77} & 10.81 \\
SoulX-FlashHead  (Lite)$^\triangle$       & 10.78 & 115.94 & 5.12 & 8.80  & -\\
SoulX-FlashHead  (Pro)$^\triangle$       & \textbf{8.31} & \textbf{103.14} & \textbf{6.04} &\textbf{ 8.46} & - \\
\bottomrule
\end{tabular}
\end{table}

%% file: new_tables/table_vfhq_results.tex
\begin{table}[t]
\caption{Quantitative comparison on the VFHQ dataset.}
\centering
\small
\setlength{\tabcolsep}{10pt}
\begin{tabular}{lccccc}
\toprule
\textbf{Method} & \textbf{FID$\downarrow$} & \textbf{FVD$\downarrow$} & \textbf{Sync-C$\uparrow$} & \textbf{Sync-D$\downarrow$} & \textbf{FPS$\uparrow$} \\
\midrule
SadTalker$^*$~\citep{zhang2023sadtalker}    & 29.80 & 191.81 & 4.49 & 8.78  & 1.60 \\
Aniportrait$^\triangle$~\citep{wei2024aniportrait}  & 36.58 & 352.94 & 1.62 & 11.73 & - \\
EchoMimic$^*$~\citep{chen2025echomimic}    & 24.69 & 193.45 & 2.93 & 10.30 & 0.79 \\
Ditto$^*$~\citep{li2024ditto}        & 27.67 & 254.05 & 3.31 & 10.26  & \underline{41.24} \\
Hallo3$^*$~\citep{cui2025hallo3}       & 23.45 & 171.00 & 4.19 & 9.60 & 0.11 \\
Sonic$^\triangle$~\citep{ji2025sonic}       & 24.03 & 142.88 & 4.64 & 8.48 & - \\
\midrule
SoulX-FlashHead  (Lite)$^*$       & 16.95 & 167.90 & 4.70 & 8.66  & \textbf{96} \\
SoulX-FlashHead  (Pro)$^*$      & \underline{14.05} & \underline{140.27} & \underline{5.53} & \underline{8.01} & 10.81 \\
SoulX-FlashHead  (Lite)$^\triangle$       & 15.18 & 156.20 & 5.33 & 8.12 & - \\
SoulX-FlashHead  (Pro)$^\triangle$       & \textbf{13.67} & \textbf{133.69} & \textbf{5.60} & \textbf{7.81} & - \\
\bottomrule
\end{tabular}
\label{table_vfhq_results}
\vspace{-1em}
\end{table}

%% file: new_figs/figure_qua.tex
\begin{figure*}[h]
    \centering
    \includegraphics[width=1.0\linewidth]{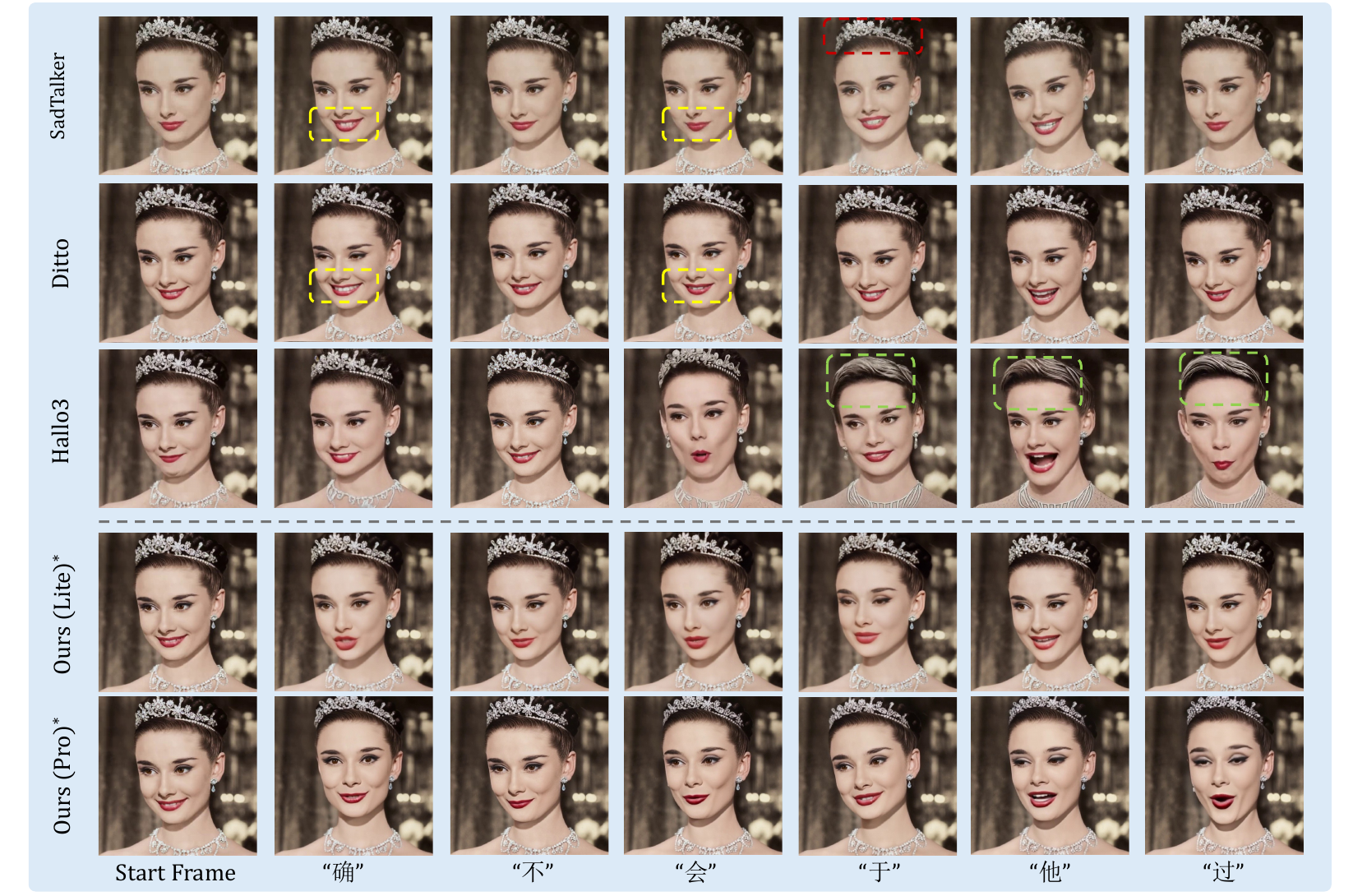}
    \caption{\textbf{Qualitative comparison on 60-second video generation at 25 fps.} \textcolor{yellow}{Yellow} dashed regions illustrate lip-synchronization mismatches in motion-based methods like Ditto and SadTalker while \textcolor{green}{green} indicators point to severe error accumulation and identity drift in Hallo3. \textcolor{red}{Red} boxes reveal holistic inconsistencies where elements like headgear separate from the subject due to the lack of unified pixel latent space modeling. In contrast, SoulX-FlashHead maintains robust lip synchronization, structural integrity, and holistic consistency throughout the sequence.}
    \label{figure_qua}
\end{figure*}

%% file: content_arxiv/5.Conclusion.tex
\section{Conclusion}
This work introduces SoulX-FlashHead which is a unified framework that effectively reconciles the conflict between high-fidelity video generation and low-latency streaming interaction. By leveraging a 1.3B-parameter DiT backbone, we implemented a robust two-stage training strategy. The Streaming-Aware Spatiotemporal Pre-training ensures stable audio-visual alignment under unstable streaming conditions while the Oracle-Guided Bidirectional Distillation significantly mitigates identity drift and error accumulation in long-sequence autoregressive generation. Extensive experiments demonstrate that our approach achieves state-of-the-art performance on benchmark datasets and offers flexible deployment options ranging from the ultra-fast 96 FPS Lite version to the high-fidelity Pro version.

Despite these advancements, the current framework exhibits limitations primarily stemming from its parameter scale. The 1.3B model possesses a constrained capacity for understanding complex physical dynamics compared to larger-scale foundational models. Consequently, our generation is primarily optimized for facial and head regions. The model currently struggles to synthesize large-amplitude body movements or intricate hand gestures with the same level of precision found in facial features. Future work will explore scaling the architecture to enhance holistic motion modeling while maintaining real-time efficiency.

%% file: content_arxiv/6.Statement.tex
\section{Ethics Statement}
This research aims to advance digital human synthesis for beneficial applications. We confirm that all datasets utilized in this study are derived from publicly accessible academic repositories. The visual demonstrations presented in this report are fully synthetic and do not contain the Personally Identifiable Information (PII) of private individuals.

We acknowledge the dual-use nature of high-fidelity video generation technology and the potential risks associated with its misuse, such as the creation of deepfakes or the spread of misinformation. We firmly condemn any malicious application of this technology and advocate for the principles of Responsible AI. To mitigate these risks, we support the development of robust forgery detection algorithms and the implementation of invisible watermarking mechanisms to ensure content transparency and traceability. We remain committed to adhering to ethical guidelines and ensuring that our contributions promote the safe and positive evolution of the field.